\documentclass[sigconf]{acmart}

\usepackage{amsmath}
\usepackage{amsthm}
\usepackage{algorithm}

\usepackage{algpseudocode}
\usepackage{amssymb}
\usepackage{amsfonts}
\usepackage{mathtools}
\usepackage{mathrsfs} 
\usepackage{multirow}
\usepackage{color}
\usepackage{xcolor}
\usepackage{colortbl}
\usepackage{balance}

\def\rowg{\rowcolor{gray!10}}

\AtBeginDocument{%
  }

\renewcommand\footnotetextcopyrightpermission[1]{}
\copyrightyear{2023}
\acmYear{2023}
\setcopyright{acmlicensed}\acmConference[MM '23]{Proceedings of the 31st ACM International Conference on Multimedia}{October 29-November 3, 2023}{Ottawa, ON, Canada}
\acmBooktitle{Proceedings of the 31st ACM International Conference on Multimedia (MM '23), October 29-November 3, 2023, Ottawa, ON, Canada}
\acmPrice{15.00}
\acmDOI{10.1145/3581783.3612113}
\acmISBN{979-8-4007-0108-5/23/10}

\begin{document}

\title[GCL: Gradient-Guided Contrastive Learning with Multi-Perspective Meta Labels]{GCL: Gradient-Guided Contrastive Learning for Medical Image Segmentation with Multi-Perspective Meta Labels}

\author{Yixuan Wu}
\affiliation{%
  \institution{Zhejiang University}
  \city{Hangzhou}
  \country{China}
}
\email{wyx_chloe@zju.edu.cn}

\author{Jintai Chen}
\authornote{Co-corresponding authors.}
\affiliation{%
  \institution{Zhejiang University}
  \city{Hangzhou}
  \country{China}
}
\email{jtchen721@gmail.com}

\author{Jiahuan Yan}
\affiliation{%
  \institution{Zhejiang University}
  \city{Hangzhou}
  \country{China}
}
\email{jyansir@zju.edu.cn}

\author{Yiheng Zhu}
\affiliation{%
  \institution{Zhejiang University}
  \city{Hangzhou}
  \country{China}
}
\email{zhuyiheng2020@zju.edu.cn}

\author{Danny Z. Chen}
\affiliation{%
  \institution{University of Notre Dame}
  \country{United States}}
\email{dchen@nd.edu}

\author{Jian Wu$^*$}
\affiliation{%
  \institution{Zhejiang University}
  \city{Hangzhou}
  \country{China}
}
\email{wujian2000@zju.edu.cn}

\renewcommand{\shortauthors}{Yixuan Wu al.}

\begin{abstract}
Since annotating medical images for segmentation tasks commonly incurs expensive costs, it is highly desirable to design an annotation-efficient method to alleviate the annotation burden. Recently, contrastive learning has exhibited a great potential in learning robust representations to boost downstream tasks with limited labels. In medical imaging scenarios, ready-made meta labels (i.e., specific attribute information of medical images) inherently reveal semantic relationships among images, which have been used to define positive pairs in previous work. However, the multi-perspective semantics revealed by various meta labels are usually incompatible and can incur intractable ``semantic contradiction" when combining different meta labels. In this paper, we tackle the issue of ``semantic contradiction" in a gradient-guided manner using our proposed \textit{Gradient Mitigator} method, which systematically unifies multi-perspective meta labels to enable a pre-trained model to attain a better high-level semantic recognition ability. Moreover, we emphasize that the fine-grained discrimination ability is vital for segmentation-oriented pre-training, and develop a novel method called \textit{Gradient Filter} to dynamically screen pixel pairs with the most discriminating power based on the magnitude of gradients. Comprehensive experiments on four medical image segmentation datasets verify that our new method GCL: (1) learns informative image representations and considerably boosts segmentation performance with limited labels, and (2) shows promising generalizability on out-of-distribution datasets.
\end{abstract}

\keywords{
medical pre-training; multi-perspective meta labels; optimization
}
\begin{teaserfigure}
\centering
  \includegraphics[width=0.85\textwidth]{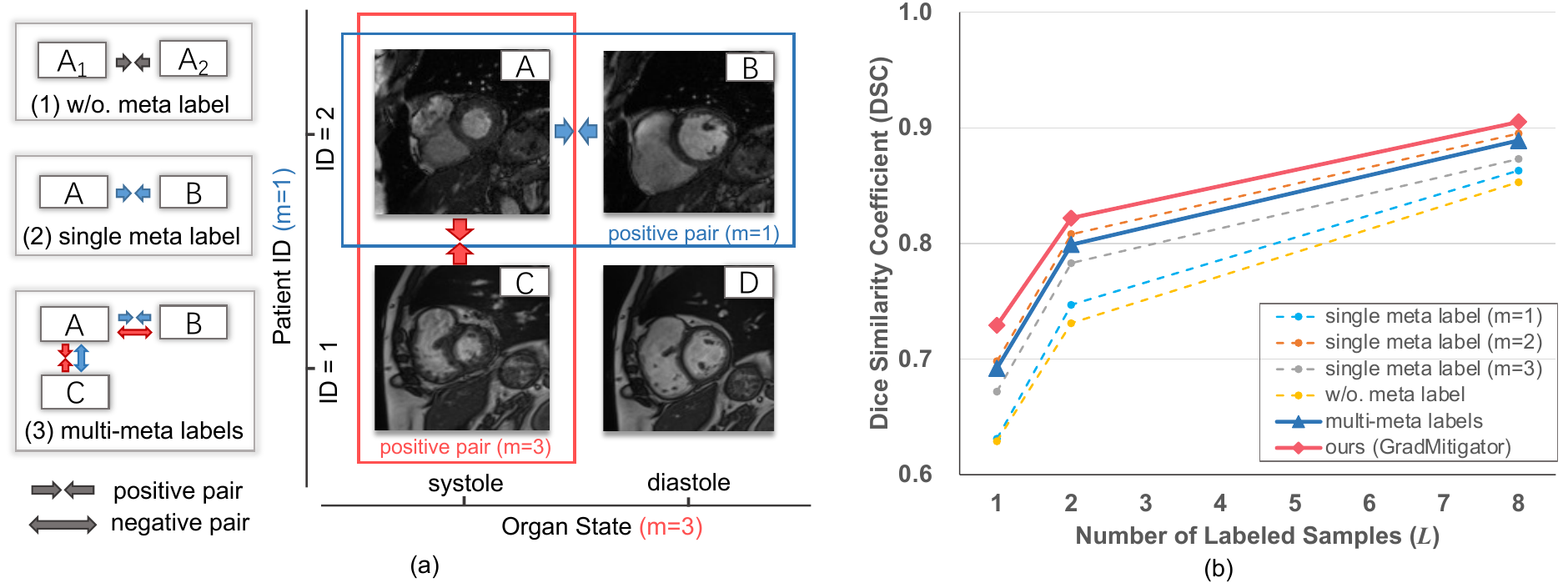}
  \caption{Illustrating the ``semantic contradiction'' problem and its negative effect. (a) Three types of common contrastive learning formulations in medical imaging scenarios: (1) vanilla contrastive learning in which a positive pair is constructed from two augmented versions ($A_{1},A_{2}$) of one image $A$; (2) a single meta label is used to define additional positive pairs, where images with an identical meta label $m$ are taken as positive pairs; (3) multiple meta labels are leveraged simultaneously to define positive pairs, in which ``semantic contradiction" may occur (e.g., images $A$ and $B$ are regarded as both a \textcolor{blue}{positive pair} and a \textcolor{red}{negative pair} simultaneously based on meta labels \textcolor{blue}{$m$=1} and \textcolor{red}{$m$=3}, respectively). Our novel gradient-guided method GradMitigator mitigates such contradiction. (b) Our preliminary experiments show that directly using multi-perspective meta labels without any additional processing can lead to worse performance (see the \textcolor{blue}{blue solid line}). Our proposed GradMitigator enables to unify and accumulate positive effects of multi-perspective meta labels (see the \textcolor{red}{red solid line}).}
\label{fig:head}
\end{teaserfigure}

\maketitle

\section{Introduction}
Cutting-edge medical image segmentation methods usually follow the paradigm of deep learning (DL) based semantic segmentation with a pixel-wise classification process. In this paradigm, pixel-wise annotation is still a big bottleneck due to the labor-intensive and time-consuming burden on medical experts. Moreover, the semantic class of each pixel is predicted independently and pixel correlation is not explicitly specified, and thus a large amount of annotations may be needed to train a comparable model~\cite{maskformer}. 

To reduce the reliance on labeled data, in this paper, we focus on contrastive learning to exploit underlying information of unlabeled data and facilitate informative model initialization for medical image segmentation with limited labels. For better segmentation, in model pre-training, we empower the model with not only recognition ability of high-level semantics (i.e., semantic similarity across the dataset) but also fine-grained discrimination ability for pixel-wise correlation.

For recognition ability, ready-made meta labels (e.g., $\mathtt{Patient\_ID}$, $\mathtt{Organ\_state}$) -- specific attribute information of different images -- are inherently a good source for models to identify semantic similarities between images and learn high-level semantics across a dataset. It was shown~\cite{glcl} that by leveraging the meta labels of slice positions as auxiliary information, contrastive learning could gain more clues to define additional positive pairs. It was verified~\cite{medaug} that the underlying pathology contained in meta labels helps learn image representations in pre-training. However, existing work focused only on utilizing a single meta label while the relationships between different meta labels were not systematically considered and the effects of them were not effectively unified.

When combining multi-perspective meta labels, a natural idea is to treat each meta label independently and sum up the effects of different meta labels directly. But, we observe in preliminary experiments (e.g., see Fig.~\ref{fig:head}(b)) that combining multiple meta labels without any additional processing may result in worse performance than using a single meta label ($m$=2). Based on this observation, we formulate the ``semantic contradiction" problem which is caused by incompatible semantics revealed by different meta labels. For example, as shown in Fig.~\ref{fig:head}(a), images $A$ and $B$ both are from the same patient but present different organ states (i.e., $A_\mathtt{Patient\_ID}\!=\!B_\mathtt{Patient\_ID}$, $A_\mathtt{Organ\_state}\!\neq\! B_\mathtt{Organ\_state}$). Inspired by multi-objective optimization theory~\cite{mgda,gradsur,gradvin,chen2018gradnorm}, we hypothesize that the contradictory semantics revealed by different meta labels can lead to divergence of model optimization and also to inferior pre-trained representations. In this work, we tackle the issue of ``semantic contradiction" in a gradient-guided manner using our proposed \textbf{Grad}ient \textbf{Mitigator} (GradMitigator), a gradient modifying method that systematically unifies positive effects of various meta labels and hence improves the model's optimization trajectory.

On the other hand, for a better pixel-level discrimination ability, the pre-training should be conducted to distinguish pixel-wise correlation for better detail-aware representations. Deviating from the common practice of pre-defining sub-image positive pairs based on physical coordinates or additional annotations, we utilize high-level semantics (i.e., image-wise semantic similarities) to first initialize a pool of positives for reserving potential positive pixels, from which optimal positives are dynamically screened to update the model with our proposed \textbf{Grad}ient \textbf{Filter} (GradFilter) method. Specifically, we define \textit{uncertainty} and \textit{hardness} as two sampling criteria, which are both characterized based on the magnitude of gradients. In this way, it is only the reliable and discriminating pixel pairs that are included for optimizing the pre-trained model.

Based on the above key components, we develop a new overall method GCL (Gradient-guided Contrastive Learning).
The main contributions of this work are as follows.
\begin{itemize}
\item 
We exploit multi-perspective meta labels to empower the model with recognition ability for high-level semantics, by mitigating the ``semantic contradiction" between meta labels in a gradient-guided fashion.
\item We extend the operating granularity of pre-training to the pixel level, where pixel-wise correlation is utilized to increase the model's fine-grained discrimination ability. Specifically, we develop a new GradFilter method to dynamically screen discriminating pixel pairs.
\item Our experiments on various medical image segmentation tasks show that, by focusing on both high-level semantics and fine-grained details, our GCL method effectively reduces the downstream model's reliance on labeled data and outperforms known related methods.
\end{itemize}

\section{Related Works}
\subsection{Contrastive Learning}
Contrastive learning was first proposed as an instance discrimination task~\cite{instance}, which aims to learn a representation space where similar instances (e.g., images) are pulled closer and different instances are pushed away. In~\cite{simclr,moco,mocov2,bt,byol,chuang2020debiased,wang2022contrastive}, a positive pair was constructed by two augmented versions of one image using transformations (e.g., crop, blur, and color transformations), while a negative pair was constructed by any two different images. To construct suitable positive and negative pairs for better representation learning, some recent work explored optimal combinations of transformations~\cite{goodview,viewmaker,wang2022contrastive} for positive pairs, while other work designed interesting sampling~\cite{negsample1} or generating~\cite{negsample2} strategies for negative pairs. In~\cite{alignment}, alignment and uniformity were identified as two key properties relevant to contrastive learning, and considerable work sought to optimize these two properties. However, mainstream contrastive formulations share two common drawbacks: (1) the criterion for positives and negatives is one-sided, which ignores semantic relationships between images, resulting in models' poor recognition ability for high-level semantics across the dataset; (2) the contrasting granularity is usually restricted to the image level while pixel correlations are overlooked, leading to inferior fine-grained discriminating ability of pre-trained models. 

\subsection{Contrastive Learning for Medical Data}
Prior work tried to use domain-specific knowledge to construct better image representations when applying contrastive learning to medical scenarios~\cite{you2022momentum,radio1,radio3,glcl,spl,medaug,chaitanya2023local,chen2021electrocardio,you2022bootstrapping,wang2022uncertainty,pgl,sscl,you2022simcvd,you2023rethinking,you2022mine}. Radiomics features were exploited as knowledge-augmentation to construct additional positive pairs for abnormality classification and localization in chest X-ray images~\cite{radio1}. In~\cite{glcl}, it was shown that by leveraging 2D slice positions, contrastive learning based pre-training could gain more clues to define additional positive pairs and the encoded image representations performed better on downstream tasks. The importance of individual images was dynamically adapted in the contrastive loss to boost performance~\cite{spl}. In~\cite{medaug}, it verified that the underlying pathology contained in meta labels helps learn pre-trained representations, and also compared the effects of different meta labels. However, the utilized domain-specific knowledge focuses only on specific features of medical datasets in a one-sided manner, while information from different perspectives is not systematically combined to characterize the overall dataset.

\subsection{Pixel-wise Contrastive Learning}
Some recent work realized that image-wise contrastive learning is classification-oriented, and extended the operating granularity from the original image level to sub-image level. Different ways to define positive pairs have been proposed. In~\cite{glcl}, the same pixel entity after different augmentations is used to form positive pairs. Spatial transformations were leveraged as a prior to deduce location relations between two augmented views, and then matched pixel pairs were formed~\cite{pgl}. In~\cite{propa}, positive pixel pairs were selected based on spatial proximity of physical coordinates. In~\cite{pixaug}, an information-guided
pixel augmentation strategy was proposed to achieve unsupervised local feature matching.
Besides, fully-supervised~\cite{cross} and semi-supervised~\cite{pixclsemisup,you2022momentum,you2022bootstrapping,wang2022uncertainty,liu2021bootstrapping,alonso2021semi,chaitanya2023local,hu2021region} settings were considered respectively, and external ground truth labels were utilized to construct positive pairs.

\section{Methodology}
\subsection{Contrastive Learning with Meta Labels}\label{sec:pre}
Given a mini-batch of $N$ unlabeled 2D images, contrastive learning aims to learn a feature extractor $f(\cdot)$ and a projection head $h(\cdot)$ to yield an image-wise representation $z\!=\!h(f(x))$ for each 2D image $x$, by pulling the representations of similar image pairs (i.e., positive pairs) together. 
In vanilla contrastive learning~\cite{simclr}, only two augmented versions of an image are regarded as a positive pair, while any two different images are taken as a negative pair even if they are semantically similar. 

In order to empower a pre-trained model with recognition ability for high-level semantics, inspired by~\cite{glcl}, we leverage the pre-specified meta labels of medical images to define additional positive pairs. \textbf{Note that such meta labels are given for free during the acquisition process of medical datasets, which reveal specific attribute information of the images} (see Sec.~\ref{subsec-details} for illustrations). 
Specifically, assume that each 2D image $x_{i}$ has $M$ kinds of meta labels (e.g., $\mathtt{Patient\_ID}$,\ $\mathtt{Organ\_state}$), denoted as $y_{i}^{m} \in \{1,\dots,C_{m}\}$, where $C_{m}$ is the class number of the meta label $m \in \{1,\dots,M\}$. Correspondingly, the image-wise contrastive loss guided by meta label $m$ is defined as:
\begin{equation}\label{equ:imgcl}
    {\mathcal{L}^{m}_\text{img}}= - \frac{1}{|\mathcal{P}^{m}_{i}|} \sum_{j \in \mathcal{P}^{m}_{i}} \log \frac{\exp \left(z_{i}\cdot z_{j} / \tau\right)}{\sum_{a=1}^{2N}\textbf{I}_{i\neq a} \exp \left(z_{i}\cdotp z_{a} / \tau\right)},
\end{equation}
where $z_{i}$ and $z_{j}$ are the representations of the anchor image and its positive respectively in the image-wise representation space ($z_{i}\!=\!h_\text{img}(f(x_{i}))$, $h_\text{img}$ projects features to image-wise representation space), $\mathcal{P}^{m}_{i}$ is the set of indices $j$ of positives $z_{j}$, $z_{a}$ denotes each representation of all augmented images in the current mini-batch except the anchor itself (including positives and negatives), and $\tau$ is a temperature parameter.

Thus, the positives come from two sources: (1) the augmented versions of the same image; (2) the different images that have the same class of the meta label $m$ considered.

\begin{figure}[t]
\centering
\includegraphics[width=\columnwidth]{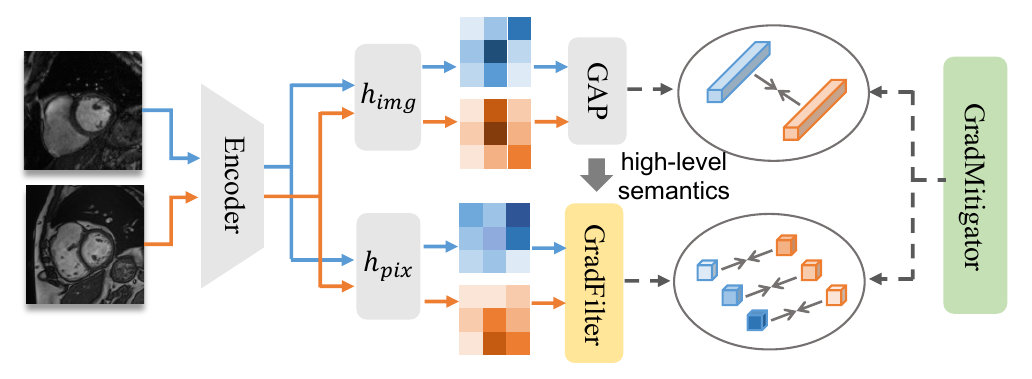}
\caption{The pipeline of GCL. 
For two images sharing the same meta label $m$, 
image head $h_{img}$ and pixel head $h_{pix}$ adopt the output from the same encoder as input and project it to their own representation spaces. 
Image-wise branch employs global average pooling (GAP) to get global features and contrasts them to learn the high-level semantics. 
GradFilter in pixel-wise branch utilizes the learned high-level semantics from image-wise branch and dynamically screens discriminating pixel pairs.
GradMitigator is applied for both image-wise and pixel-wise contrasts to alleviate conflicts between different meta labels (only one meta label is illustrated in figure for simplicity). 
}
\label{fig:pip}
\end{figure}

\subsection{Gradient Mitigator}\label{sec:gradkiller}
To combine the multi-perspective meta labels, a direct way is to simply sum up the contrastive losses guided by all the $M$ meta labels, training jointly to minimize the average loss and update model parameters $\theta$ by:
\begin{equation}\label{equ:mtl}
    {\theta^{*}=\mathop{\arg\min}_{\theta \in \textbf{R}} (\frac{1}{M} \sum_{m=1}^{M} \mathcal{L}^{m}(\theta))}.
\end{equation}
But, this can incur the ``semantic contradiction" issue since high-level semantics revealed by different meta labels may be incompatible. In Fig.~\ref{fig:head}(a), images $A$ and $B$ are taken from the same patient but present different organ states, and thus are regarded as a positive pair and a negative pair in computing contrastive losses guided by $\mathtt{Patient\_ID}$ and $\mathtt{Organ\_state}$, respectively. We hypothesize that such contradiction can lead to divergence of model optimization and also to inferior pre-trained representations. Poor performance when combining multiple meta labels without any additional processing in preliminary experiments (see Fig.~\ref{fig:head}(b)) verifies our hypothesis. 

\textbf{We aim to tackle the issue of ``semantic contradiction" in a gradient-guided manner,} given the learning process of DL networks is dictated by gradients with respect to network parameters ($\theta$) -- usually back-propagated in the network during gradient descent~\cite{gradient1,gradient2}. 
Specifically, let \resizebox{.24\linewidth}{!}{$\mathbf{g}_{m}\!=\!\nabla_{\theta} \mathcal{L}^{m}(\theta)$} denote an individual gradient guided by meta label $m$, and \resizebox{.39\linewidth}{!}{$\mathbf{g}\!=\!\nabla_{\theta} \mathcal{L}(\theta)\!=\!\frac{1}{M} \sum_{m=1}^{M} \mathbf{g}_{m}$} be the average gradient. With a learning rate \resizebox{.21\linewidth}{!}{$\alpha$, $\theta \!\gets\! \theta-\alpha\mathbf{g}$} gives the steepest descent update when optimizing Eq.~(\ref{equ:mtl}). However, if the individual gradient $\mathbf{g}_{m}$ conflicts with $\mathbf{g}$, following Eq.~(\ref{equ:mtl}) directly will interfere the optimization trajectory guided by meta label $m$.

\begin{figure}[t]
\centering
\includegraphics[width=1\columnwidth]{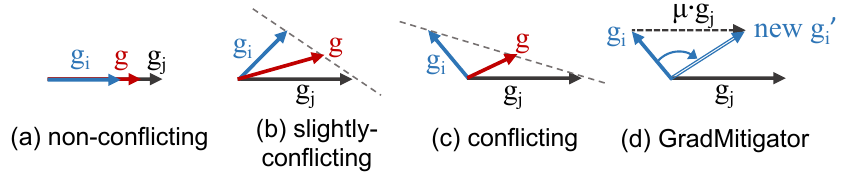}
\caption{Diagrams for illustrating conflicting gradients and our proposed GradMitigator method. \textcolor{blue}{Blue}, black, and \textcolor{red}{red} arrows represent meta-level gradients \textcolor{blue}{$\mathbf{g}_{i}$}, $\mathbf{g}_{j}$, and averaged gradient \textcolor{red}{$\mathbf{g}$}, respectively. (a)-(c) Three types of gradient relationships. (d) Our GradMitigator method aims to modify gradient \textcolor{blue}{$\mathbf{g}_{i}$} by injecting a weighted component of $\mathbf{g}_{j}$ to mitigate the gradient interference.}
\label{fig:arrow}
\end{figure}

Thus, we propose the novel GradMitigator method to mitigate the gradient interference by modifying conflicting gradients of different meta-level contrastive losses. As shown in Fig.~\ref{fig:arrow}, we study three types of gradient relationships based on cosine similarity $\omega_{ij}$ between meta-level gradients $\mathbf{g}_{i}$ and $\mathbf{g}_{j}$: (a) non-conflicting (i.e., $\omega_{ij}\!=\!1$); (b) slightly-conflicting (i.e., $0\!\leq\!\omega_{ij}\!<\!1$); (c) conflicting (i.e., $\omega_{ij}\!<\!0$). 

\begin{algorithm}
  \caption{The updating process with Gradient Mitigator}\label{alg:gradkiller}
  \begin{algorithmic}[1]
  \Require Model parameters $\theta$, meta labels $m \in \{1,\dots,M\}$, loss functions $\mathcal{L}^{m}(\theta)$, and EMA weight $\beta$
  \State Initialize time-step $t=0$, EMA variable $\hat{\omega}^{(0)}_{ij}=0 ,\forall i,j$ 
  \State Compute $\mathbf{g}_{m} \gets \nabla_{\theta}\mathcal{L}^{m}(\theta), \forall m$
    \For{$i \in \{1,\dots,M\} \ $}
    \State Set $\mathbf{g}'_{i} \gets \mathbf{g}_{i}$
      \For{$j \in \{1,\dots,M\} \;\backslash \;\{i\} \ $}
        \State Compute $\omega^{(t)}_{ij} \gets \frac{\mathbf{g}_{i}'\cdot \mathbf{g}_{j}}{\Vert \mathbf{g}_{i}'\Vert \Vert \mathbf{g}_{j} \Vert}$
        \State Update $\hat{\omega}^{(t)}_{ij}\gets (1-\beta)\hat{\omega}^{(t-1)}_{ij}+\beta\omega^{(t)}_{ij}$
        \If{$\omega^{(t)}_{ij} < \hat{\omega}^{(t)}_{ij} $}
        \State \resizebox{0.8\linewidth}{!}{$\!\mathbf{g}_{i}'\!=\!\mathbf{g}_{i}'\!+\!\frac{\|\mathbf{g}_{i}^{\prime}\|
          (\hat{\omega}_{ij}^{(t)} \!\sqrt{1-\left(\omega_{i j}^{(t)}\right)^{2}}
          \!-\!\omega_{i j}^{(t)} \!\sqrt{1-\left(\hat{\omega}_{i j}^{(t)}\right)^{2}})}
          {\left\|\mathbf{g}_{j}\right\| \sqrt{1-\left(\hat{\omega}_{i j}^{(t)}\right)^{2}}}
          \!\cdot\! \mathbf{g}_{j}$}
        \EndIf 
      \EndFor
    \EndFor
    \State Update $\Delta \theta \gets \mathbf{g}'= \frac{1}{M}\sum_{i=1}^{M}\mathbf{g}'_{i}$ 
    \State Update time-step $t \gets t+1$
  \end{algorithmic}
\end{algorithm}

\textbf{The goal of GradMitigator is to softly eliminate conflicting components of gradients, seeking agreement between the individual meta-level gradients. }
Alg.~\ref{alg:gradkiller} presents the updating process. We first initialize the target cosine similarity $\hat{\omega}^{(0)}_{ij}$ as $0$, and pre-compute all the gradients $\mathbf{g}_{m}$ of contrastive functions $\mathcal{L}^{m}(\theta)$ guided by different meta labels $m$. We use $i,j$ as two meta labels for illustrating an updating process: 
At the current time-step $t$, if the computed cosine similarity between two gradients, i.e., $\omega^{(t)}_{ij}=\frac{\mathbf{g}_{i}\cdot \mathbf{g}_{j}}{\Vert \mathbf{g}_{i}\Vert \Vert \mathbf{g}_{j} \Vert}$, is smaller than the target value $\hat{\omega}^{(t)}_{ij}$, we modify one gradient $\mathbf{g}_{i}$ by injecting a weighted component of the other gradient $\mathbf{g}_{j}$, i.e., $\mathbf{g}_{i}'=\mathbf{g}_{i}+\mu\cdot\mathbf{g}_{j}$, such that the resulting cosine similarity softly matches the target value $\hat{\omega}^{(t)}_{ij}$. Based on the Law of Sines, the weight $\mu$ is computed.
This modifying process is described as:
\begin{equation}\label{equ:update_g}
    \resizebox{.85\linewidth}{!}{$\displaystyle
    \mathbf{g}'_{i}\!=\!\mathbf{g}_{i}\!+\!
          \frac{\|\mathbf{g}_{i}\|
          (\hat{\omega}_{ij}^{(t)} \!\sqrt{1-\left(\omega_{i j}^{(t)}\right)^{2}}
          \!-\!\omega_{i j}^{(t)} \!\sqrt{1-\left(\hat{\omega}_{i j}^{(t)}\right)^{2}})}
          {\left\|\mathbf{g}_{j}\right\| \sqrt{1-\left(\hat{\omega}_{i j}^{(t)}\right)^{2}}}
          \!\cdot\! \mathbf{g}_{j}$}.
\end{equation}
Note that the value of the target cosine similarity is not fixed. Instead, we use the exponential moving average (EMA) for: (1) avoiding drastic change of the target value during training, and (2) bootstrapping for a potentially better target value in a self-adapting manner. This is why we call it `softly', as:
\begin{equation}\label{equ:ema}
        {\hat{\omega}^{(t)}_{ij} = (1-\beta)\hat{\omega}^{(t-1)}_{ij}+\beta\omega^{(t)}_{ij}}.
\end{equation}
In this manner, a new average gradient $\mathbf{g}'$ is obtained to update the model parameters $\theta$ in practice.

\subsection{Gradient Filter}\label{sec:gradfilter}
It is important to note that the image-level contrastive learning mentioned above is classification task-oriented, which often uses a pooling-like operation to aggregate features from all spatial locations to obtain an image-wise representation. In such a situation, pixel correlation is not explicitly concerned, restricting the pre-trained model's fine-grained discrimination ability, especially for segmentation tasks. In this work, we extend the operating granularity to the pixel level. Existing work usually defines pixel-wise positive pairs based on: (i) the same pixel entity (after different augmentations), (ii) corresponding physical coordinates, and (iii) additional ground truth labels. However, (i) restricts the source of positive pairs to the same image; (ii) largely relies on the assumption that different images in the dataset are well aligned and registered; (3) requires a large number of pixel-wise annotations. All these are not practical in real medical scenarios. 

Instead, we propose to utilize the learned high-level semantics between images to pre-define a pool of positives, and dynamically introduce optimal positives from this pool to update the model with our devised Gradient Filter method. 

First, for each pixel, we build its pool of positives based on pixel affinity. The pixel affinity $\mathcal{A}$ is computed based on corresponding features in the image-wise representation space (before the pooling-like operation). Suppose pixel $i(u)$ in image $x_{i}$ is an anchor pixel, its Top-$K$ similar pixels $j(v)$ in image $x_{j}$ are formed as the positive pool $\mathcal{P}_{u}$ for pixel $i(u)$, and all the remaining pixels in image $x_{j}$ are regarded as negatives $\mathcal{N}_{u}$. Note that the anchor pixel and its positives/negatives are not restricted to being from the same image (after various augmentations); instead, image $x_{j}$ provides all the image-wise positives of $x_{i}$ defined in Sec.~\ref{sec:pre} guided by a meta label $m$. Our pixel-wise contrastive loss is defined as:
\begin{equation}\label{equ:pixcl}
    \resizebox{.89\linewidth}{!}{$
            \displaystyle
    {\mathcal{L}_\text{pix}^{m}}= - \frac{1}{|\mathcal{P}_{u}|}\! \sum_{u^{+} \in \mathcal{P}_{u}}\!\log\!\frac{\exp \left(u\cdot u^{+} / \tau\right)}{\exp(u\cdotp u^{+} / \tau)\!+\!\sum\limits_{u^{-} \in \mathcal{N}_{u}}\!\!\exp(u\cdotp u^{-}/ \tau)}$},
\end{equation}
\begin{equation}
    {\mathcal{A}_{i(u)j(v)}}=\frac{z_{i(u)}\cdot z_{j(v)}}{\Vert z_{i(u)} \Vert \,\Vert z_{j(v)} \Vert},
\end{equation}
where $u, u^{+}$, and $u^{-}$ denote respectively the anchor pixel, its positive and negative in the pixel-wise representation space (i.e., $u=h_\text{Pix}(f(x_{i(u)}))$, where $h_\text{Pix}$ projects features to the pixel-wise representation space), and $z_{i(u)}$ and $z_{j(v)}$ are corresponding features of pixels $i(u)$ and $j(v)$ in the image-wise representation space (e.g., $z_{i(u)}=h_\text{img}(f(x_{i(u)}))$). 

\begin{figure}[t]
\centering
\includegraphics[width=0.85\columnwidth]{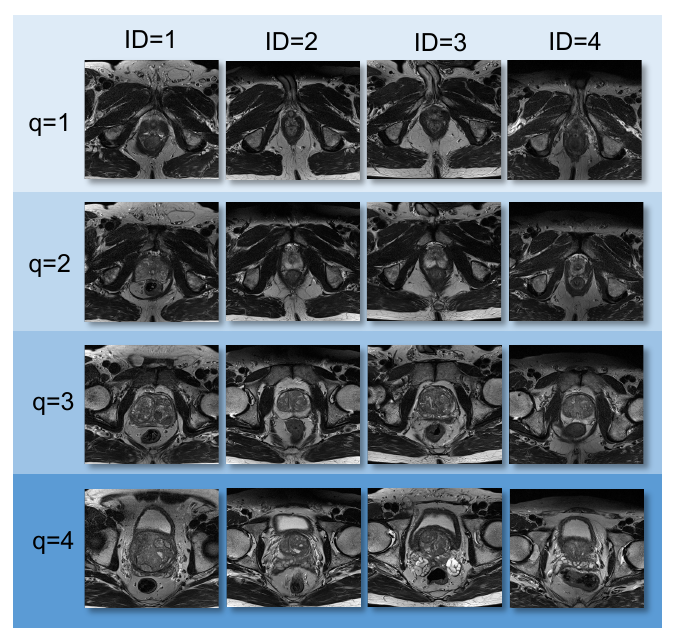}
\vskip -1.2 em
\caption{Examples of 2D slices taken from different quantiles ($q=1,2,3,4$) for four patients on Prostate dataset. One can see that the slices with the same quantile from different patients contain relatively similar anatomical structures.}
\label{fig:quantile}
\end{figure} 

Next, to further enhance the effectiveness of the defined positives, we consider that an ``ideal'' positive should both be reliable (i.e., with a low \textit{uncertainty}) for the right optimization direction of the model and have a certain degree of \textit{hardness} for constantly optimizing the model's decision boundary~\cite{negsample1}. Hence, we propose GradFilter for screening positive pixels with a high discriminating power based on two criteria, \textit{uncertainty} and \textit{hardness}, from the pre-defined positive pool to update the model. Since DL networks are optimized using gradient-based methods, we characterize these two criteria by gradient magnitudes induced by different positives. The gradient of the pixel-wise contrastive loss w.r.t. the anchor pixel representation can be described as:
\begin{equation}\label{equ:g}
    \resizebox{.89\linewidth}{!}{$
            \displaystyle
{\frac{\partial \mathcal{L}^{m}_\text{pix}}{\partial u}\!=\!-\frac{1}{\tau\!\left|\mathcal{P}_{u}\right|}\!\sum_{u^{+} \in \mathcal{P}_{u}} \!
\left(\!\frac{(u^{+}\!-\!u^{-})\!\sum\limits_{u^{-} \in \mathcal{N}_{u}}\!\exp(u\cdotp u^{-}/ \tau)}{\exp(u\cdotp u^{+} / \tau)\!+\!\sum\limits_{u^{-} \in \mathcal{N}_{u}}\!\!\exp(u\cdotp u^{-}/ \tau)}\!\right)}$}.
\end{equation}

One may see that a harder positive usually has a smaller dot product with the anchor, which brings more gradient contribution than easier positives. Conversely, we consider that the model is more certain about a positive if a smaller gradient is induced and hence a little update is performed at the current optimization direction~\cite{gradal}. With these two criteria, we aim to make reconciliation inspired by Self-Pace Learning~\cite{selfpace}, \textbf{following the learning process of humans, so that the model learns better when feeding samples from easy to hard to it. }

Specifically, at time-step $t$, for each positive $u^{+}\in\mathcal{P}_{u}$, we compute the corresponding gradient with respect to the parameters of the last layer of the encoder. A pace function $g(t)$ is defined to specify the positive pool size so that only positives with the $g(t)$ lowest gradients are actually used, as:
\begin{equation}\label{equ:gt}
    g(t)=[1+\frac{1}{4}\log(\frac{t}{T}+e^{-4})]\cdot \left|\mathcal{P}_{u}\right|,
\end{equation}
where $T$ denotes the total number of training steps. This scheme schedules how the positives are introduced to the training process: At the beginning, positives with high certainty are preferred, and as the training progresses, more harder positives are introduced.

\subsection{Training Objective}
As shown in Fig.~\ref{fig:pip}, we jointly perform image-wise contrastive learning (Sec.~\ref{sec:pre}) and pixel-wise contrastive learning (Sec.~\ref{sec:gradfilter}) under the guidance of a meta label $m$. The effects of all the $M$ multi-perspective meta labels are dynamically unified to optimize the model with modified gradients using our GradMitigator method (Sec.~\ref{sec:gradkiller}), by:
\begin{equation}
{\mathcal{L}_\text{GCL}=\sum_{m=1}^{M}\left(\mathcal{L}_\text{img}^{m}+\mathcal{L}_\text{pix}^{m}\right)}.
\end{equation}

\begin{figure}[t]
\centering
\includegraphics[width=0.85\columnwidth]{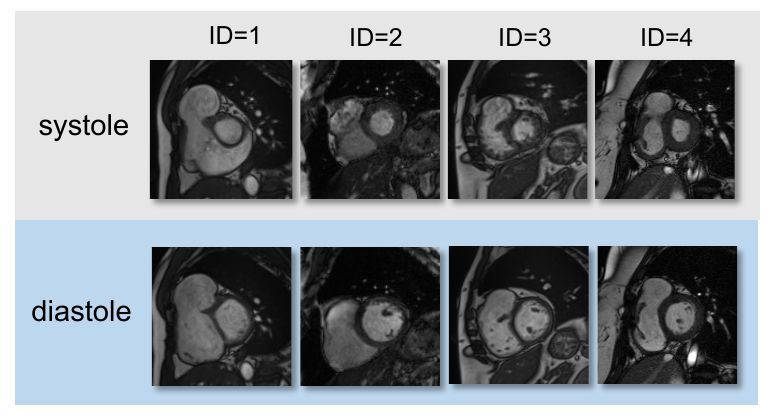}
\vskip -1.2 em
\caption{Examples of 2D slices taken in different organ states (i.e., systole and diastole) from four different patients on the ACDC dataset. It can be seen that the appearances of the heart in different states are different.}
\label{fig:state}
\end{figure} 

\begin{table*}[t]
\centering
\caption{Comparison of our proposed GCL method and related pre-training methods on four datasets in DSC performance. $L$ denotes the number of provided labeled samples in fine-tuning. The best results are marked in \textbf{bold}.}
\vskip -1 em
\label{table:pt}
\setlength{\tabcolsep}{1.2mm}
\resizebox{\textwidth}{40mm}{
\begin{tabular}{l|ccc|ccc|ccc|ccc}
\hline
\multirow{2}{*}{Method} & \multicolumn{3}{c|}{ACDC}                                                                                     & \multicolumn{3}{c|}{Prostate}                                                                                 & \multicolumn{3}{c|}{MMWHS}                                                                                    & \multicolumn{3}{c}{ACDC $\xrightarrow{}$ HVSMR}                                                                 \\\cline{2-13} 
                        & $L$=1                                & $L$=2                                & $L$=8                                 & $L$=1                                & $L$=2                                & $L$=8                                 & $L$=1                                & $L$=2                                & $L$=8                                 & $L$=2                                & $L$=4                                & $L$=6                                \\ \hline
                        \rowg \multicolumn{13}{c}{Baseline} \\\hline
Random Init.            & 0.598$_{\pm.023}$                             & 0.682$_{\pm.012}$                              & 0.847$_{\pm.008}$                               & 0.477$_{\pm.039}$                              & 0.547$_{\pm.027}$                              & 0.645$_{\pm.013}$                               & 0.443$_{\pm.017}$                              & 0.623$_{\pm.012}$                              & 0.792$_{\pm.008}$                               & 0.743$_{\pm.037}$                              & 0.817$_{\pm.025}$                              & 0.847$_{\pm.017}$                              \\ \hline
                        \rowg \multicolumn{13}{c}{Image-level Contrastive Learning} \\\hline
SimCLR                  & 0.629$_{\pm.037}$                              & 0.731$_{\pm.022}$                              & 0.853$_{\pm.017}$                               & 0.519$_{\pm.049}$                              & 0.578$_{\pm.033}$                              & 0.663$_{\pm.022}$                               & 0.489$_{\pm.022}$                              & 0.667$_{\pm.018}$                              & 0.799$_{\pm.009}$                               & 0.737$_{\pm.047}$                              & 0.807$_{\pm.022}$                              & 0.842$_{\pm.023}$                              \\
MoCo                    & 0.623$_{\pm.032}$                              & 0.723$_{\pm.024}$                              & 0.851$_{\pm.012}$                               & 0.503$_{\pm.047}$                              & 0.577$_{\pm.032}$                              & 0.660$_{\pm.027}$                               & 0.501$_{\pm.018}$                              & 0.652$_{\pm.020}$                              & 0.787$_{\pm.012}$                               & 0.733$_{\pm.048}$                              & 0.809$_{\pm.029}$                              & 0.840$_{\pm.025}$                              \\\hline
                        \rowg \multicolumn{13}{c}{Pixel-level Contrastive Learning} \\\hline
PixPro & 0.642$_{\pm.025}$          & 0.754$_{\pm.022}$          & 0.863$_{\pm.020}$          & 0.544$_{\pm.029}$          & 0.597$_{\pm.024}$          & 0.670$_{\pm.022}$          & 0.519$_{\pm.024}$          & 0.672$_{\pm.017}$          & 0.798$_{\pm.017}$   & 0.753$_{\pm.033}$                              & 0.821$_{\pm.033}$                    & 0.849$_{\pm.018}$   \\
DenseCL & 0.633$_{\pm.027}$          & 0.742$_{\pm.021}$          & 0.859$_{\pm.020}$          & 0.537$_{\pm.029}$          & 0.589$_{\pm.024}$          & 0.668$_{\pm.022}$          & 0.514$_{\pm.027}$          & 0.688$_{\pm.018}$          & 0.797$_{\pm.023}$   & 0.752$_{\pm.032}$                              & 0.817$_{\pm.037}$                    & 0.842$_{\pm.022}$   \\
PointRC & 0.647$_{\pm.023}$          & 0.760$_{\pm.017}$          & 0.867$_{\pm.013}$          & 0.552$_{\pm.028}$          & 0.601$_{\pm.022}$          & 0.673$_{\pm.018}$          & 0.524$_{\pm.025}$          & 0.677$_{\pm.010}$          & 0.799$_{\pm.014}$   & 0.750$_{\pm.038}$                              & 0.820$_{\pm.031}$                    & 0.848$_{\pm.012}$   \\
\hline
  \rowg \multicolumn{13}{c}{Medical Pre-training} \\\hline
GLCL                    & 0.702$_{\pm.020}$                              & 0.783$_{\pm.015}$                              & 0.881$_{\pm.011}$                               & 0.572$_{\pm.029}$                              & 0.612$_{\pm.018}$                              & 0.687$_{\pm.022}$                               & 0.557$_{\pm.015}$                              & 0.689$_{\pm.014}$                              & 0.801$_{\pm.005}$                               & 0.778$_{\pm.033}$                              & 0.825$_{\pm.023}$                              & 0.852$_{\pm.022}$                              \\
PosiCL                  & 0.688$_{\pm.021}$                              & 0.803$_{\pm.018}$                              & 0.886$_{\pm.012}$                               & 0.554$_{\pm.031}$                              & 0.606$_{\pm.017}$                              & 0.689$_{\pm.017}$                               & 0.533$_{\pm.012}$                              & 0.692$_{\pm.011}$                              & 0.814$_{\pm.003}$                               & 0.780$_{\pm.029}$                              & 0.827$_{\pm.020}$                              & 0.855$_{\pm.018}$                              \\
SSCL                    & 0.697$_{\pm.017}$                              & 0.785$_{\pm.011}$                              & 0.892$_{\pm.009}$                               & 0.587$_{\pm.023}$                              & 0.621$_{\pm.012}$                              & 0.692$_{\pm.009}$                               & 0.547$_{\pm.017}$                              & 0.694$_{\pm.009}$                              & 0.812$_{\pm.004}$                               & 0.779$_{\pm.030}$                              & 0.839$_{\pm.022}$                              & 0.859$_{\pm.019}$     \\
SPL  & 0.699$_{\pm.023}$          & 0.801$_{\pm.012}$          & 0.889$_{\pm.014}$          & 0.588$_{\pm.023}$          & 0.624$_{\pm.017}$          & 0.688$_{\pm.023}$          & 0.560$_{\pm.014}$          & 0.694$_{\pm.013}$          & 0.814$_{\pm.009}$      & 0.782$_{\pm.030}$                              & 0.827$_{\pm.022}$                              & 0.857$_{\pm.019}$  \\\hline
  \rowg \multicolumn{13}{c}{Pretext task Pre-training} \\\hline

Rotation                & 0.592$_{\pm.037}$                              & 0.689$_{\pm.022}$                              & 0.852$_{\pm.014}$                               & 0.508$_{\pm.043}$                              & 0.562$_{\pm.037}$                              & 0.659$_{\pm.029}$                               & 0.447$_{\pm.020}$                              & 0.635$_{\pm.019}$                              & 0.787$_{\pm.012}$                               & 0.752$_{\pm.044}$                              & 0.819$_{\pm.024}$                              & 0.847$_{\pm.022}$                              \\
Inpainting              & 0.605$_{\pm.025}$                              & 0.701$_{\pm.018}$                              & 0.863$_{\pm.005}$                               & 0.503$_{\pm.033}$                              & 0.557$_{\pm.020}$                              & 0.668$_{\pm.013}$                               & 0.463$_{\pm.018}$                              & 0.650$_{\pm.010}$                              & 0.792$_{\pm.008}$                               & 0.757$_{\pm.023}$                              & 0.820$_{\pm.017}$                              & 0.849$_{\pm.009}$                              \\
Jigsaw                  & 0.595$_{\pm.028}$                              & 0.712$_{\pm.015}$                              & 0.874$_{\pm.010}$                               & 0.501$_{\pm.038}$                              & 0.570$_{\pm.023}$                              & 0.667$_{\pm.012}$                               & 0.437$_{\pm.044}$                              & 0.642$_{\pm.012}$                              & 0.793$_{\pm.007}$                               & 0.748$_{\pm.038}$                              & 0.823$_{\pm.020}$                              & 0.852$_{\pm.012}$                              \\\hline

GCL (ours)             & \textbf{0.729$_{\pm.014}$ }         & \textbf{0.812$_{\pm.009}$}          & \textbf{0.905$_{\pm.007}$ }         & \textbf{0.606$_{\pm.019}$ }         & \textbf{0.631$_{\pm.013}$ }         & \textbf{0.701$_{\pm.012}$ }         & \textbf{0.572$_{\pm.010}$}          & \textbf{0.709$_{\pm.008}$}          & \textbf{0.819$_{\pm.005}$ }         & \textbf{0.797$_{\pm.022}$ }         & \textbf{0.849$_{\pm.017}$}          & \textbf{0.870$_{\pm.007}$}          \\ \hline

\end{tabular}}
\end{table*}

\section{Experiments}
\subsection{Experimental Setup}
We conduct four sets of experiments, investigating: (1) the informativeness of learned representations compared with other pre-training methods; (2) the effectiveness to reduce downstream task's reliance on labeled data compared with other semi-supervised methods; (3) the generalizability on out-of-distribution datasets; and (4) the effects of each designed components.

\subsection{Implementations}
\label{subsec-details}

\noindent {\bf Training and Evaluation.}
Note that our proposed GCL is a pre-training method. Following~\cite{glcl,spl,sscl}, the performance of our GCL method is evaluated in a ``pre-training and fine-tuning" paradigm: (1) GCL method is used to pre-train a U-Net encoder; and (2) the pre-trained weights are regarded as initialization for the downstream segmentation network to be fine-tuned with limited labels. The performance of our method is indicated by the segmentation accuracy. All the fine-tuning experiments are repeated $5$ times.  Segmentation results are reported in the form of mean (standard deviation) with the dice similarity coefficient (DSC).

\noindent {\bf Data.}
We evaluate the performance of our GCL method on four public medical image datasets: ACDC~\cite{acdc}, Prostate~\cite{prostate}, MMWHS~\cite{mmwhs,mmwhs2}, and HVSMR~\cite{hvsmr}. These four datasets have different anatomical structures (i.e., cardiac, prostate), modalities (i.e., MRI, CT), resolutions, and sizes, allowing comprehensive evaluation of our method. For ACDC, we leverage the meta labels of $\mathtt{Patient\_ID}$ ($m$=1), $\mathtt{Slice\_quantile}$ ($m$=2) (i.e., the quantile of a 2D image along one axis), and $\mathtt{Organ\_state}$  ($m$=3) (i.e., systole or diastole). For the other datasets, the meta labels of $\mathtt{Patient\_ID}$ and $\mathtt{Slice\_quantile}$ are used. 
Each dataset is split into a pre-training set $X_{ptr}$ and a fine-tuning set $X_{ft}$; the fine-tuning set $X_{ft}$ is further split into a training set $X_{tr}$, a validation set $X_{val}$, and a test set $X_{ts}$. We use $X_{ptr}$ to pre-train the GCL model without ground truth labels, and use $X_{ft}$ to fine-tune the pre-trained encoder on the downstream task and report segmentation performance. A small number of samples in $X_{tr}$ are randomly chosen as labeled samples (e.g., $L=1,2,8$). 

\noindent {\bf Illustration of Meta Labels.}
We show 2D images (or slices) with their meta labels in Figs.~\ref{fig:quantile} and~\ref{fig:state}, in order to provide intuitive illustrations of the usage of different meta labels. $\mathtt{Slice\_quantile}$ represents the quantile of a 2D image/slice along one axis (i.e., the $z$-axis for the Prostate, MMWHS, and HVSMR datasets, and the short axis for the ACDC dataset). Thus, all the slices in a 3D image are divided into four parts, and the corresponding quantile is indicated by $q$ ($q\!=\!1,2,3,4$). In Fig.~\ref{fig:quantile}, we illustrate slices taken from different quantiles, for four different patients. One can see that the slices with the same quantile from different patients contain similar anatomical structures. Besides, $\mathtt{Organ\_state}$ indicates the state of a target organ at the time of scanning (e.g., the systole or diastole state of the heart). It can be seen from Fig.~\ref{fig:state} that the appearances of the heart in different states show considerable differences.

\noindent {\bf Model Details.}
Our contrastive formulation follows~\cite{simclr}. The architecture of the encoder follows U-Net~\cite{unet}. The two projection heads $h_\text{img}$ and $h_\text{pix}$ share the same design, which consists of $1\times 1$ convolution, ReLU, and $1\times 1$ convolution. The hidden layer’s dimension of the projection head is $512$, keeping the same as its input channels, and the final output dimension is $128$, the same as~\cite{simclr}. In the fine-tuning stage, we employ U-Net as our segmentation network.

\begin{table*}[t]
\centering
\caption{Comparison of our proposed GCL method and semi-supervised methods on three datasets with limited labeled data provided in DSC performance. $L$ denotes the number of provided labeled samples. The best results are marked in \textbf{bold}.}
\vskip -1 em
\label{table:semi}
\setlength{\tabcolsep}{2mm}
\resizebox{\textwidth}{18mm}{
\begin{tabular}{l|ccc|ccc|ccc}
\hline
\multirow{2}{*}{Method} & \multicolumn{3}{c|}{ACDC}                                                                                     & \multicolumn{3}{c|}{Prostate}                                                                                 & \multicolumn{3}{c}{MMWHS}                                                                                                                \\\cline{2-10} 
                        & $L$=1                                & $L$=2                                & $L$=8                                 & $L$=1                                & $L$=2                                & $L$=8                                 & $L$=1                                & $L$=2                                & $L$=8                                                               \\ \hline
                        
Baseline           & 0.598$_{\pm.023}$                             & 0.682$_{\pm.012}$                              & 0.847$_{\pm.008}$                               & 0.477$_{\pm.039}$                              & 0.547$_{\pm.027}$                              & 0.645$_{\pm.013}$                               & 0.443$_{\pm.017}$                              & 0.623$_{\pm.012}$                              & 0.792$_{\pm.008}$                                                         \\ \hline

Adv. Training           & 0.662$_{\pm.012}$                              & 0.749$_{\pm.013}$                              & 0.849$_{\pm.007}$                                & 0.544$_{\pm.023}$                              & 0.587$_{\pm.020}$                              & 0.681$_{\pm.010}$                              & 0.531$_{\pm.018}$                              & 0.679$_{\pm.012}$                              & 0.790$_{\pm.007}$                                                         \\
Mean Teacher            & 0.674$_{\pm.012}$                              & 0.771$_{\pm.007}$                              & 0.857$_{\pm.004}$                              & 0.526$_{\pm.014}$                              & 0.557$_{\pm.008}$                              & 0.657$_{\pm.004}$                                & 0.538$_{\pm.012}$                              & 0.692$_{\pm.008}$                              & 0.809$_{\pm.004}$                                                        \\
Mixup                   & 0.667$_{\pm.009}$                              & 0.773$_{\pm.010}$                              & 0.862$_{\pm.005}$                               & 0.531$_{\pm.017}$                              & 0.598$_{\pm.012}$                              & 0.677$_{\pm.008}$                               & 0.549$_{\pm.014}$                              & 0.683$_{\pm.013}$                              & 0.797$_{\pm.009}$                                                          \\\hline
GCL (ours)             & 0.729$_{\pm.014}$          & 0.812$_{\pm.009}$          & 0.905$_{\pm.007}$          & 0.606$_{\pm.019}$          & 0.631$_{\pm.013}$          & 0.701$_{\pm.012}$          & 0.572$_{\pm.010}$          & 0.709$_{\pm.008}$          & 0.819$_{\pm.005}$                 \\
Ours + Mixup              & 0.754$_{\pm.007}$          & 0.833$_{\pm.008}$          & 0.911$_{\pm.002}$          & 0.612$_{\pm.012}$          & 0.633$_{\pm.014}$          & \textbf{0.713$_{\pm.002}$} & 0.621$_{\pm.007}$          & 0.722$_{\pm.008}$          & 0.822$_{\pm.003}$                \\
Ours + M.T.                 & \textbf{0.762$_{\pm.010}$} & \textbf{0.842$_{\pm.008}$} & \textbf{0.913$_{\pm.002}$} & \textbf{0.614$_{\pm.014}$} & \textbf{0.643$_{\pm.008}$} & 0.711$_{\pm.005}$          & \textbf{0.628$_{\pm.008}$} & \textbf{0.731$_{\pm.005}$} & \textbf{0.828$_{\pm.005}$} \\ \hline
Fully Superv.        & \multicolumn{3}{c|}{0.914$_{\pm.003}$ ($L$=50)}                                                                                    & \multicolumn{3}{c|}{0.703$_{\pm.005}$ ($L$=18)}                                                                                    & \multicolumn{3}{c}{0.812$_{\pm.009}$ ($L$=10)}                                                                                                                                            \\\hline
\end{tabular}}
\end{table*}

\begin{table*}[t]
\centering
\caption{Ablation study with different components of our GCL method on three datasets in DSC performance. $L$ denotes the number of labeled samples in fine-tuning. The best results are marked in \textbf{bold}.}
\vskip -1 em
\label{table:abl}
\setlength{\tabcolsep}{1.2mm}
\resizebox{\textwidth}{23.4mm}{
\begin{tabular}{l|ccc|ccc|ccc}
\hline
\multirow{2}{*}{Method}     & \multicolumn{3}{c|}{ACDC}                        & \multicolumn{3}{c|}{Prostate}                    & \multicolumn{3}{c}{MMWHS}                        \\ \cline{2-10} 
                            & $L$=1            & $L$=2            & $L$=8            & $L$=1            & $L$=2            & $L$=8            & $L$=1            & $L$=2            & $L$=8            \\ \hline
Base & 0.629$_{\pm.037}$                              & 0.731$_{\pm.022}$                              & 0.853$_{\pm.017}$                               & 0.519$_{\pm.049}$                              & 0.578$_{\pm.033}$                              & 0.663$_{\pm.022}$                               & 0.489$_{\pm.022}$                              & 0.667$_{\pm.018}$                              & 0.799$_{\pm.009}$                                                            \\\hline
Base + single meta label ($m$=$1$) & 0.631$_{\pm.024}$          & 0.747$_{\pm.019}$          & 0.863$_{\pm.018}$          & 0.527$_{\pm.032}$          & 0.589$_{\pm.020}$          & 0.678$_{\pm.022}$          & 0.529$_{\pm.020}$          & 0.684$_{\pm.012}$          & 0.799$_{\pm.002}$          \\
Base + single meta label ($m$=$2$)   & 0.688$_{\pm.020}$          & 0.784$_{\pm.013}$          & 0.880$_{\pm.012}$          & 0.573$_{\pm.027}$          & 0.617$_{\pm.019}$          & 0.688$_{\pm.024}$          & 0.549$_{\pm.013}$          & 0.683$_{\pm.014}$          & 0.801$_{\pm.007}$          \\
Base + single meta label ($m$=$3$)      & 0.672$_{\pm.018}$          & 0.783$_{\pm.018}$          & 0.873$_{\pm.009}$          & \multicolumn{3}{c|}{/}                           & \multicolumn{3}{c}{/}                            \\
Base + multi-meta labels & 0.681$_{\pm.022}$          & 0.784$_{\pm.020}$          & 0.878$_{\pm.017}$          & 0.574$_{\pm.023}$          & 0.614$_{\pm.018}$          & 0.685$_{\pm.012}$          & 0.548$_{\pm.023}$          & 0.684$_{\pm.021}$          & 0.802$_{\pm.014}$          \\
Base + multi-meta labels + GradMiti.  & 0.709$_{\pm.019}$          & 0.804$_{\pm.018}$          & 0.894$_{\pm.012}$          & 0.592$_{\pm.030}$          & 0.623$_{\pm.028}$          & 0.692$_{\pm.014}$          & 0.559$_{\pm.017}$          & 0.697$_{\pm.013}$          & 0.811$_{\pm.008}$          \\
\hline
Base + PixCL               & 0.635$_{\pm.024}$          & 0.747$_{\pm.022}$          & 0.860$_{\pm.018}$          & 0.545$_{\pm.028}$          & 0.593$_{\pm.027}$          & 0.663$_{\pm.019}$          & 0.513$_{\pm.022}$          & 0.685$_{\pm.017}$          & 0.793$_{\pm.014}$          \\    
Base + PixCL + GradFilter  & 0.652$_{\pm.020}$          & 0.775$_{\pm.019}$          & 0.871$_{\pm.014}$          & 0.562$_{\pm.024}$          & 0.612$_{\pm.028}$          & 0.679$_{\pm.017}$          & 0.536$_{\pm.019}$          & 0.694$_{\pm.016}$          & 0.803$_{\pm.009}$          \\\hline
Full model (frozen)                 & 0.709$_{\pm.011}$         & 0.797$_{\pm.007}$         & 0.882$_{\pm.006}$         & 0.593$_{\pm.014}$          & 0.617$_{\pm.012}$          & 0.684$_{\pm.008}$          & 0.563$_{\pm.011}$         & 0.694$_{\pm.008}$          & 0.809$_{\pm.006}$          \\
Full model (ours)                 & 0.729$_{\pm.014}$         & 0.812$_{\pm.009}$         & 0.905$_{\pm.007}$         & 0.606$_{\pm.019}$          & 0.631$_{\pm.013}$          & 0.701$_{\pm.012}$          & 0.572$_{\pm.010}$         & 0.709$_{\pm.008}$          & 0.819$_{\pm.005}$          \\
Full model (merged data)                 & \textbf{0.734$_{\pm.012}$}          & \textbf{0.817$_{\pm.009}$}          & \textbf{0.908$_{\pm.004}$}          & \textbf{0.614$_{\pm.014}$ }         & \textbf{0.634$_{\pm.016}$}          & \textbf{0.702$_{\pm.009}$}          & \textbf{0.579$_{\pm.008}$}          & \textbf{0.714$_{\pm.007}$}          & \textbf{0.822$_{\pm.005}$}          \\

\hline
\end{tabular}}
\end{table*}

\noindent
{\bf Training Details.}
The GCL pre-training is performed on four NVIDIA GeForce RTX 3090 GPUs. We train with the SGD optimizer~\cite{sgd} for $300$ epochs, and the cosine learning rate scheduler is adopted, with a batch size of $48$ and a learning rate of $0.1$. In the fine-tuning stage, we train the segmentation network with limited labels for $300$ epochs. The Adam optimizer~\cite{adam} and cosine learning rate scheduler are used, with a batch size of $5$ and a learning rate of $10^{-4}$. The temperature $\tau$ is set to $0.1$ following~\cite{simclr}. $K$ is set to $0.3$ when defining the positive pool. The EMA weight $\beta$ is set to $10^{-2}$. All the parameters $\theta$ of the encoder are updated individually based on the modified gradients when applying the proposed GradMitigator method.

\subsection{Comparison with Pre-training Methods}
To evaluate the informativeness of learned image representations, we compare our GCL with several groups of pre-training methods. (1) Image-level contrastive learning, including SimCLR~\cite{simclr}, MoCo~\cite{moco}. (2) Pixel-level contrastive learning, including PixPro~\cite{propa}, DenseCL~\cite{densecl}, and PointRC~\cite{prc}. (3) Medical pre-training, including GLCL~\cite{glcl}, PosiCL~\cite{posicl}, SSCL~\cite{sscl}, and SPL~\cite{spl}. (4) Pretext task pre-training, including Rotation~\cite{rotate}, Inpainting~\cite{inpaint}, and Jigsaw~\cite{jigsaw}. 

\noindent
{\bf Results.} 
Table~\ref{table:pt} summarizes the results on four downstream segmentation tasks, which are used to indicate the pre-training performance. $L$ denotes the number of provided labeled samples in fine-tuning. 
As baselines, we train the downstream segmentation network with random initialization (train from scratch). 
One can see that the pretext task pre-training (i.e., Rotation, Inpainting, and Jigsaw) provides less informative initialization, which shows worse performance when labeled data is extremely limited (i.e., $L=1,2$). 
Besides, the general image-level contrastive learning methods (i.e., SimCLR and MoCo) provide useful initialization to some extent. 
And when extending the contrasting granularity to the pixel level (i.e., PixPro, DenseCL, and PointRC), the pre-training performance gets further boosted.
In addition, the pre-training methods designed in medical scenarios (i.e., GLCL, PosiCL, SSCL, and SPL) perform better than those general pre-training methods, suggesting that single-source domain-specific information in medical images provides useful clues to some extent.
Yet, our GCL still boosts the downstream segmentation accuracy to a large extent (e.g., $0.131$, $0.129$, $0.129$, and $0.054$ in DSC on the ACDC, Prostate, MMWHS, and HVSMR dataset, when $L$=1, respectively). This is because our GCL method effectively unifies the information from multi-perspective meta labels. 
Beyond that, it can be seen that the fewer labeled samples (i.e., $L=1,2$) provided in fine-tuning, the more significant the superiority of our GCL pre-training method. This validates the necessity of pre-training for segmentation in medical scenarios where limited labeled data can be provided.

\begin{figure*}[t]
\centering
\includegraphics[width=1.54\columnwidth]{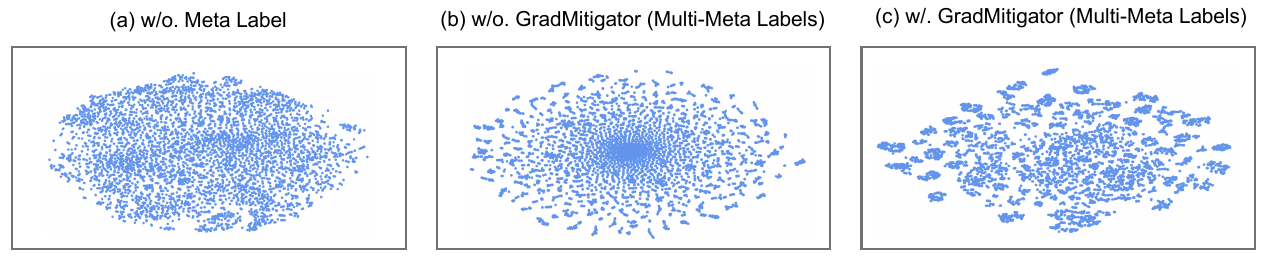}
\caption{Some t-SNE visualization examples of the learned representations on the ACDC dataset. (a) No meta label is used, and the pre-trained model is degraded to the vanilla contrastive learning, where the learned representations are evenly distributed. (b) Multiple meta labels are simultaneously used without any additional processing, where the learned representations are excessively diffused. (c) With our proposed GradMitigator method, obvious semantic separation is formed.}
\label{fig:vis_nocolor}
\end{figure*} 

\subsection{Comparison with Semi-supervised Methods}
To further investigate the effectiveness of our GCL to reduce downstream task's reliance on labeled data, we compare it with semi-supervised methods, including Adv.~Training~\cite{adv}, Mean Teacher~\cite{mt}, and Mixup~\cite{mixup}. 

\noindent
{\bf Results.}
In Table~\ref{table:semi}, the baseline is trained from scratch without other designs, and the segmentation results under full supervision (i.e., all labeled samples in the training set are provided). The improvements of semi-supervised methods (i.e., Adv.~Training, Mean Teacher, and Mixup) are limited and largely depend on specific datasets. For example, Adv.~Training performs worse on the ACDC dataset but performs better on the Prostate dataset. In contrast, our GCL method performs well on all the datasets (e.g., $0.131$, $0.129$, $0.129$, and $0.054$ in DSC on the ACDC, Prostate, MMWHS, and HVSMR dataset, when $L$=1, respectively).
Moreover, it can be seen that our method shows promising compatibility with semi-supervised methods (e.g., Mixup and Mean Teacher), leading to further performance gains (e.g., $0.164$, $0.137$, and $0.185$ in DSC on the ACDC, Prostate, and MMWHS dataset, when $L$=1, respectively), and even surpassing the results under full supervision on the Prostate and MMWHS datasets.

\subsection{Generalizability}
To further examine whether the image representations learned by our GCL method have good generalizability, we conduct pre-training and fine-tuning on different datasets. Specifically, our proposed GCL is used to pre-train an encoder on ACDC dataset, and the downstream segmentation network is fine-tuned on HVSMR.

\noindent
{\bf Results.}
The last column of Table~\ref{table:pt} presents the segmentation performance on the out-of-distribution dataset. The image-level and pixel-level contrastive learning (i.e., SimCLR, MoCo, PixPro, DenseCL, and PointRC) and pretext task pre-training (i.e., Rotation, Inpainting, and Jigsaw) give little performance gain for the downstream task. In contrast, our GCL method shows superiority compared with the other pre-training methods. This is partially due to the multi-perspective meta labels that are unified to empower the model with recognition ability for high-level semantics, which are commonly found in different medical image datasets.

\subsection{Ablation Study}
The results of ablation study are presented in  Table~\ref{table:abl}. The base model refers to the results that all designs in our GCL method are removed, with vanilla image-level contrastive learning retained.

\noindent
{\bf Effects of Introducing Meta Labels.}
We compare the performance pre-trained with different meta labels one-by-one. The meta labels of $\mathtt{Patient\_ID}$ ($m$=$1$), $\mathtt{Slice\_quantile}$ ($m$=$2$), and $\mathtt{Organ\_state}$ ($m$=$3$, only for the ACDC dataset) are included. One can see that by introducing the meta labels as additional clues, the segmentation performance gets well improved. In particular, the meta label of $\mathtt{Slice\_quantile}$ ($m$=$2$) provides more gains on all the datasets.

\noindent
{\bf Effects of Our Gradient Mitigator.}
We further investigate the effect of simultaneously using multiple meta labels. It can be seen that when combine all meta labels without any additional processing, 
the performance is poor and is even worse than using a single meta label ($m$=$2$). This verifies our hypothesis that the ``semantic contradiction'' between different meta labels can incur divergence of model optimization. 
Further, only by modifying the conflicting gradients with our proposed GradMitigator can multi-perspective meta labels be unified to synergistically optimize the pre-training process. 
Meanwhile, in Fig.~\ref{fig:vis_nocolor}, we utilize t-SNE~\cite{tsne} to visualize the learned representations on the ACDC dataset. In Fig.~\ref{fig:vis_nocolor}(b), when introducing multiple meta labels without any additional processing, the learned representations are excessively diffused. Instead, in Fig.~\ref{fig:vis_nocolor}(c), when applying our proposed GradMitigator, the obvious semantic separation is formed, suggesting the effectiveness of our GCL on exploring the semantic similarity between images and thus capturing high-level semantics across the dataset.

\noindent
{\bf Effects of Pixel-wise Contrastive Learning and Gradient Filter.}
To explore the effects of fine-grained contrasting granularity, we add the pixel-wise contrastive learning (PixCL) component to our base model, where the GradFilter strategy is not applied first (remove the pace function $g(t)$ (\textit{cf.} Eq.~(\ref{equ:gt})), and use all the positives in the pre-defined positive pool to update the model). It can be seen that the PixCL boosts the downstream performance to some extent, indicating that fine-grained contrast is necessary to segmentation-oriented pre-training, which do contribute to the model's recognition ability for segmentation details. 
Moreover, when further adding our proposed GradFilter method, the downstream segmentation accuracy gets largely boosted. Therefore, by considering both \textit{uncertainty} and \textit{hardness} to screen optimal pixel-wise positives, our pixel-wise contrastive learning exert its substantial effectiveness.

\noindent
{\bf The scope of pre-training set.} 
To further explore the extensibility of our GCL, we merge the pre-training sets of all used datasets (i.e., ACDC, Prostate, and MMWHS) to pre-train the encoder with our full GCL model. The results are shown in the last row of Table~\ref{table:abl} (denoted as ``merged data"). Compared with using the single dataset to pre-train (denoted as ``ours"), incorporating more pre-training data from other datasets (even from different organs and modalities) does contribute to the pre-training performance.

\noindent
{\bf Effects of fine-tuning the pre-trained parameters.}
During the fine-tuning process, we freeze the entire pre-trained encoder and only fine-tune the decoder. The results in Table~\ref{table:abl} suggest that the pre-trained encoder has certain feature extraction capability, while fine-tuning can further enhance the performance. Moreover, when an extremely limited number of labeled samples ($L$=1) is provided, our GCL pre-training method plays a more significant role.

\section{Conclusions}
In this paper, we proposed to systematically unify multi-perspective meta labels without incurring the ``semantic contradiction'' issue by modifying their corresponding gradients. Further, when extending the contrast granularity to the pixel level, our new Gradient Filter method helps dynamically screen positive pixel pairs with the most discriminating power. Compared to other contrastive formulations, our method empowers the pre-trained model with both recognition ability for high-level semantics and discrimination ability for pixel-wise correlation in a gradient-guided manner. Extensive experiments on four public datasets verified that our GCL method not only learns informative image representations for downstream segmentation with extremely limited labels, but also shows promising generalizability on out-of-distribution datasets.

\bibliographystyle{ACM-Reference-Format}
\bibliography{ref}

\end{document}